\documentclass{article}

\usepackage[nonatbib,preprint]{neurips_2023}

\usepackage[utf8]{inputenc} % allow utf-8 input
\usepackage[T1]{fontenc}    % use 8-bit T1 fonts
\usepackage{hyperref}       % hyperlinks
\usepackage{url}            % simple URL typesetting
\usepackage{booktabs}       % professional-quality tables
\usepackage{amsfonts}       % blackboard math symbols
\usepackage{nicefrac}       % compact symbols for 1/2, etc.
\usepackage{microtype}      % microtypography
\usepackage{xcolor}         % colors
\usepackage[accsupp]{axessibility} % Improves PDF readability for those with disabilities.

\usepackage{floatrow}
\newfloatcommand{capbtabbox}{table}[][\FBwidth]
\usepackage{booktabs}
\usepackage{multirow}
\usepackage{adjustbox}
\usepackage{transparent}
\usepackage{caption}
\usepackage{subcaption}

\usepackage{float}
\floatstyle{plaintop}
\restylefloat{table}

\definecolor{rebuttalcolor}{RGB}{0,165,135}

\title{Exploring SAM Ablations for Enhancing Medical Segmentation in Radiology and Pathology}

\author{%
  Amin Ranem\\
  \And
  Niklas Babendererde\\
  \And
  Moritz Fuchs\\
  \And
  Anirban \\\textbf{Mukhopadhyay}\\
}

\begin{document}

\maketitle

\begin{abstract}
Medical imaging plays a critical role in the diagnosis and treatment planning of various medical conditions, with radiology and pathology heavily reliant on precise image segmentation. The “Segment Anything Model” (SAM) has emerged as a promising framework for addressing segmentation challenges across different domains.
In this white paper, we delve into SAM, breaking down its fundamental components and uncovering the intricate interactions between them. We also explore the fine-tuning of SAM and assess its profound impact on the accuracy and reliability of segmentation results, focusing on applications in radiology (specifically, brain tumor segmentation) and pathology (specifically, breast cancer segmentation).
Through a series of carefully designed experiments, we analyze SAM's potential application in the field of medical imaging. We aim to bridge the gap between advanced segmentation techniques and the demanding requirements of healthcare, shedding light on SAM's transformative capabilities.
\end{abstract}

\section{Introduction}
Medical imaging is the keystone of modern healthcare, helping with diagnosis and treatment planning of a wide spectrum of diseases. Its effectiveness is based on the precise delineation of anatomical structures and pathological regions within medical images. Achieving this accuracy is particularly paramount in fields such as radiology and pathology, where critical decisions affecting patient outcomes are heavily based on \emph{semantic image segmentation} \cite{jones2018, smith2020}.

Image segmentation involves partitioning an image into meaningful regions or objects, a process that has historically posed challenges due to the inherent complexity and variability of medical imagery. Conventional methods have made significant strides in this regard, yet they \emph{often fall short when faced with the challenges of radiological and pathological data} \cite{brown2019, wang2017}.

The recently published \textit{Segment Anything Model} (SAM), is a framework that has gained attention for its potential to revolutionize image segmentation across various domains \cite{kirillov2023segment}. SAM goes beyond the boundaries of traditional approaches, by expecting a specific prompt of samples or bounding boxes, to offer a versatile and promising solution to the existing challenges of accurate segmentation.

In this white paper, we examine SAM in-depth, exploring its inner workings to uncover the core components and understand how they interact. We investigate how fine-tuning SAM affects the precision and dependability of segmentation, with a special focus on its use in two critical medical imaging fields: radiology, where we emphasize \emph{brain tumor segmentation} \cite{menze2014multimodal, bakas2017advancing, bakas2018identifying, bakas10segmentation}, and pathology, with a focus on \emph{breast cancer segmentation} \cite{bcss}.

Through a strategically structured series of experiments and analyses based on recent publications around SAM, our goal is to discover SAM's full potential in the field of medical imaging. We seek to connect existing work and their insights with the rigorous demands of healthcare, providing insights into how SAM can transform medical diagnosis.

\section{Understanding SAM Components}
    The SAM model takes a raw image as input and passes it through the image encoder, which outputs the image embeddings.
    Querying these embeddings is more efficient. After combining them with the convoluted mask, they pass through the mask decoder together with the output from the prompt encoder.
    
    The prompt encoder can take sparse prompts such as points, boxes, text or dense prompts like masks as input. When combining its output with the combined mask and image embeddings, the mask decoder maps the mask as demanded from the prompt to the segmented output image.
    
    The SAM model uses \emph{focal loss}~\cite{lin2017focal} combined with \emph{dice loss}~\cite{milletari2016v} on the resulting predicted masks.
    Medical applications can benefit from the resulting flexibility of SAM: The image encoder can process various images from different domains, such as pathology or radiology. Moreover, the prompt encoder allows different types of input, which allows clinicians to choose the most suitable one.
    
\section{Fine-Tuning SAM for Medical Imaging}
\begin{figure}[]
    \centering
    \includegraphics[width=1.0\textwidth]{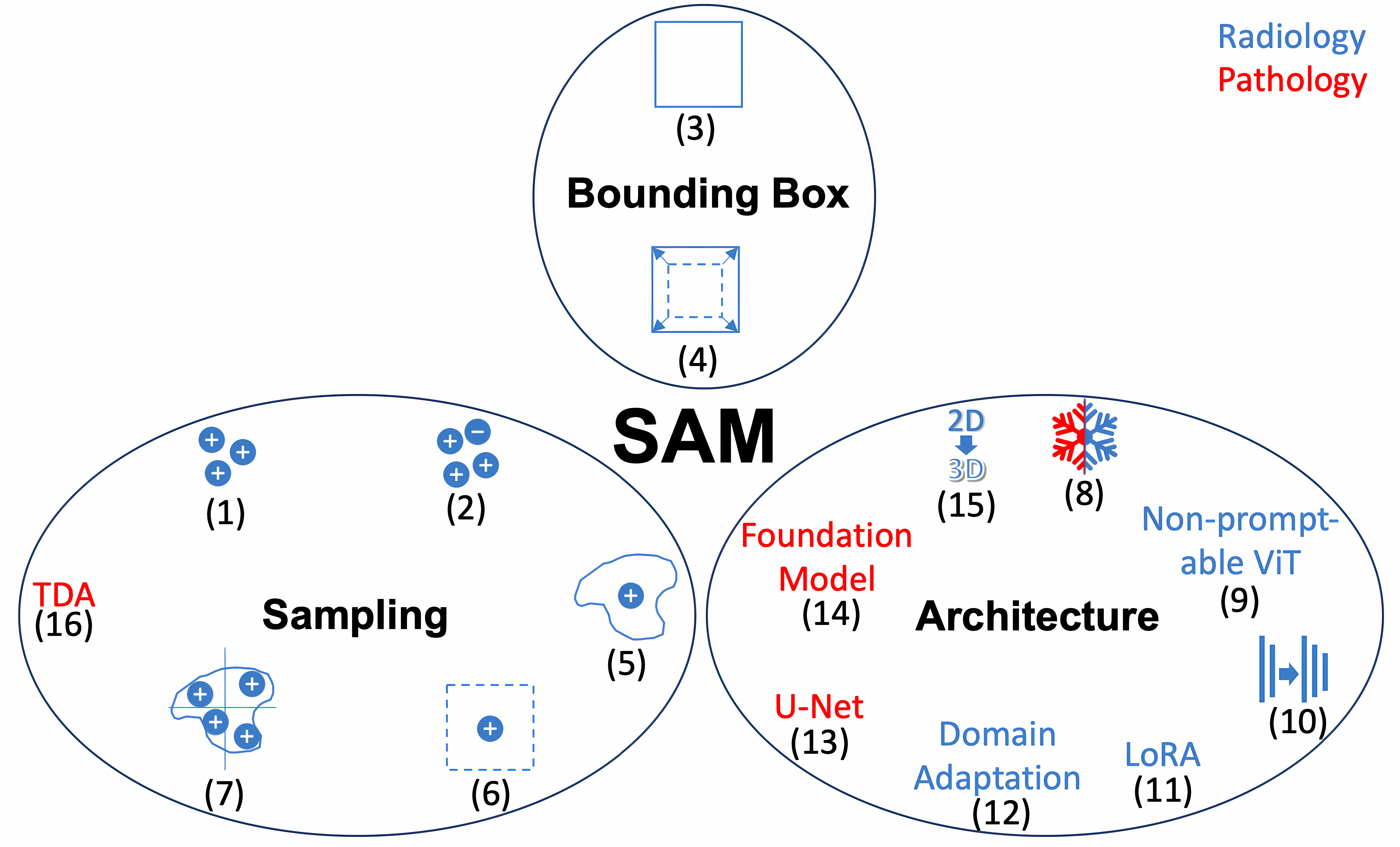}
    \caption{Overview of the given modifications of SAM for radiology (blue) and pathology (red). We cluster them into three groups: Changes to the sampling strategies, architectural changes and modifications to the bounding box of the prompt.}
    \label{fig:sam_cluster}
\end{figure}
Using SAM in new data domains requires fine-tuning with given images and corresponding ground truth segmentation masks.
For example, when a medical professional wants to apply SAM to a new type of image data such as brain MRIs, the complete model needs fine-tuning on these.
If we want to segment whole slide images for prostate tumors, we need to repeat this process.

As we are evaluating on different datasets, we also need to retrain but due to limited time and resources, we freeze SAM, except the segmentation head.
Only training the segmentation head decreases the resulting segmentation performance, but as we show in the results, it's sufficient and decreases the training time, allowing us to conduct more experiments.

Table~\ref{tab:references} and figure~\ref{fig:sam_cluster} give an overview of the different types of architectures and sampling strategies that exist for medical images. We focus on the sampling strategies and bounding boxes and train the segmentation head with them for 75 epochs. To evaluate the resulting segmentation head, we generate segmentations from the test set and compare it with the ground truth. 
The section “Training Setup and Metrics” provides more details on this process and the hyperparameters.
As we are evaluating the performance on using different prompt types, we train the segmentation head on them and then evaluate the resulting model.

\begin{table}[]
\resizebox{1.0\columnwidth}{!}{%
\begin{tabular}{|c|c|c|c|c|}
\hline
\textbf{Method}                                                                                            & \textbf{Category} & \textbf{Literature}                                                                                                                    & \textbf{ID} & \textbf{Implemented} \\ \hline
Using a Bounding Box for Input                                                                             & Bounding Box      & \cite{cheng2023sam}                                                                                                                           & 1           & Yes                  \\ \hline
Adding Jitter to the Bounding Box Location                                                                 & Bounding Box      & \cite{de2023zero}                                                                                                                             & 2           & Yes                  \\ \hline
Changing the Number of positive Points                                                                     & Sampling          & \cite{cheng2023sam}                                                                                                                           & 3           & Yes                  \\ \hline
Adding negative points                                                                                     & Sampling          & \cite{cheng2023sam}                                                                                                                           & 4           & Yes                  \\ \hline
Using only the Centroid of the Ground Truth Mask                                                           & Sampling          & \cite{de2023zero}                                                                                                                             & 5           & Yes                  \\ \hline
Using the Center of the curated Bounding Box                                                               & Sampling          & \cite{cheng2023sam}                                                                                                                           & 6           & Yes                  \\ \hline
\begin{tabular}[c]{@{}c@{}}Dividing GT Mask into 4 and sampling \\ one random Point from each\end{tabular} & Sampling          & \cite{de2023zero}                                                                                                                             & 7           & Yes                  \\ \hline
\begin{tabular}[c]{@{}c@{}}Fine-tuning SAM and keeping the \\ Image Encoder Frozen\end{tabular}            & Architecture      & \begin{tabular}[c]{@{}c@{}}\cite{hu2019attention}, \cite{ma2021medsam}, \\ \cite{li2021sam}, \cite{zhang2022segment},\\ \cite{hu2023efficiently}, \cite{cui2023all}\end{tabular} & 8           & Yes                  \\ \hline
\begin{tabular}[c]{@{}c@{}}Making ViT non-promtable by \\ changing the Prediction Head\end{tabular}        & Architecture      & \cite{hu2023efficiently}                                                                                                                      & 9           & No                   \\ \hline
Changing Depth in the CNN Prediction Head                                                                  & Architecture      & \cite{hu2023efficiently}                                                                                                                      & 10          & No                   \\ \hline
Introduce LoRA in the Image Encoder                                                                        & Architecture      & \cite{zhang2023customized}                                                                                                                    & 11          & No                   \\ \hline
Domain-Specific Adaptations                                                                                & Architecture      & \cite{wu2023medical}                                                                                                                          & 12          & No                   \\ \hline
\begin{tabular}[c]{@{}c@{}}Using SAM for training conventional \\ Segmentation Model\end{tabular}          & Architecture      & \cite{shaharabany2023autosam}, \cite{zhang2023input}                                                                                                 & 13          & No                   \\ \hline
\begin{tabular}[c]{@{}c@{}}Adding Pathology Foundation Model as \\ additional Feature Encoder\end{tabular} & Architecture      & \cite{zhang2023sam}                                                                                                                           & 14          & No                   \\ \hline
Modifying ViT from 2D to 3D                                                                                & Architecture      & \cite{gong20233dsam}, \cite{glatt2023topological}                                                                                                    & 15          & No                   \\ \hline
Using TDA for guiding Prompt                                                                               & Architecture      & \cite{glatt2023topological}                                                                                                                   & 17          & No                   \\ \hline 
\end{tabular}%
\caption{References to the modifications of SAM for radiology and pathology, their cluster categories, IDs in figure~\ref{fig:sam_cluster} and implementation status for the evaluation in this work.}
\label{tab:references}
}
\end{table}

\section{Experimental Setup}
In the following section, we provide details of our setup including the datasets, hardware that is used for training and specific training configuration and metrics for the evaluation. In addition, we give examples of the sampling strategies for the bounding boxes and illustrate them with images.

\subsection{BraTS and BCSS}
\begin{figure}[htbp]
  \centering
  \begin{minipage}{0.45\textwidth}
    \centering
    \begin{subfigure}{0.45\textwidth}
      \includegraphics[width=\textwidth]{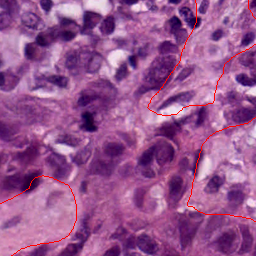}
      \caption{}
    \end{subfigure}
    \hfill
    \begin{subfigure}{0.45\textwidth}
      \includegraphics[width=\textwidth]{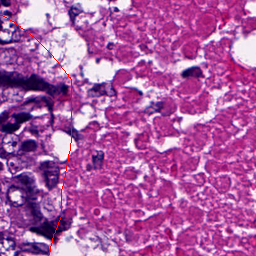}
      \caption{}
    \end{subfigure}

    \begin{subfigure}{0.45\textwidth}
      \includegraphics[width=\textwidth]{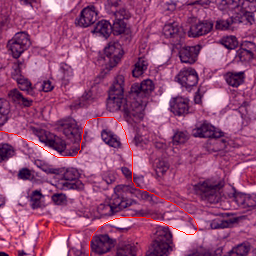}
      \caption{}
    \end{subfigure}
    \hfill
    \begin{subfigure}{0.45\textwidth}
      \includegraphics[width=\textwidth]{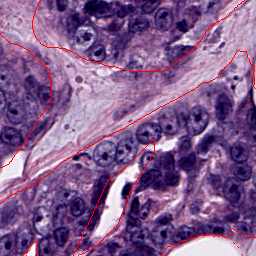}
      \caption{}
    \end{subfigure}
  \end{minipage}%
  \hfill
  \begin{minipage}{0.45\textwidth}
    \centering
    \begin{subfigure}{0.45\textwidth}
      \includegraphics[width=\textwidth]{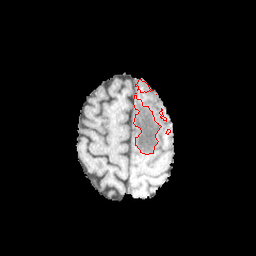}
      \caption{}
    \end{subfigure}
    \hfill
    \begin{subfigure}{0.45\textwidth}
      \includegraphics[width=\textwidth]{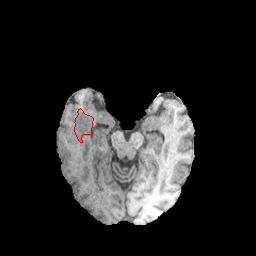}
      \caption{}
    \end{subfigure}

    \begin{subfigure}{0.45\textwidth}
      \includegraphics[width=\textwidth]{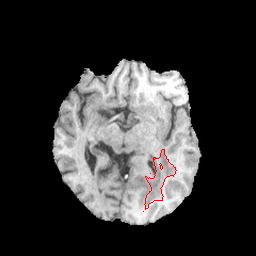}
      \caption{}
    \end{subfigure}
    \hfill
    \begin{subfigure}{0.45\textwidth}
      \includegraphics[width=\textwidth]{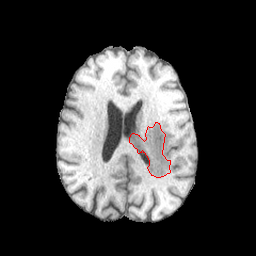}
      \caption{}
    \end{subfigure}
  \end{minipage}
  \caption{Samples patches from BCSS (a-d) and BraTS 2020 (e-h) with their respective tumor ground truths marked in red.}
  \label{fig:sample_patches}
\end{figure}
As the focus of this work is on possible applications of SAM in the medical domains, we evaluate on the radiology dataset BraTS and the pathology dataset BCSS. Figure~\ref{fig:sample_patches} shows sample patches from both datasets. The following sections provide more details on each dataset.
\subsubsection{BCSS}
The \emph{Breast Cancer Semantic Segmentation (BCSS)}~\cite{bcss} dataset contains annotated whole slide images of breast cancer tissue regions from TCGA.
For this experiment, 4400 patches with a resolution of 256$\times$256 pixels are used for training and 1100 patches for testing, leading to 80\% train data and 20\% test data.

All patches are fully annotated with the labels tumor, stroma, inflammatory, necrosis and other.
To achieve a binary segmentation, this work focuses on the tumor by setting it to the label 1 and all other labels to the combined label 0.
We pass the resulting segmentation masks through a frozen SAM model for extracting the form embeddings.

\subsubsection{BraTS 2020}
The radiology dataset \emph{Brain Tumor Segmentation (BraTS) 2020} \cite{brats1, brats2, brats3} contains MRI scans of patients with brain tumors and includes complete segmentation masks for tumor segmentation.
Similarly to the pathology dataset, 80\% of the data are used for training and 20\% for testing and all labels, except tumor, are set to 0 to achieve a binary segmentation task. The patch size is also the same as in BCSS with 256$\times$256 pixels, and we extract the form embeddings from a pass of the segmentation masks through the same frozen SAM model.

\subsubsection{Preprocessing}
To get the form embeddings for the given ground truth segmentation masks, they are passed through a frozen SAM model. We use the implementation of MedSAM\footnote{https://github.com/bowang-lab/MedSAM/blob/main/pre\textunderscore MR.py} to achieve this. To save time during training and as we freeze the SAM model, we only train the segmentation head.

\subsection{Hardware Setup}
Two different systems are used for the evaluation:
Both use a 16 Core CPU with 64 GB of DDR4 RAM and a 2 TB NVMe SSD.
The first one uses an RTX 4090 with 24 GB of VRAM for training and the second one has an RTX 3090 with also 24 GB of RAM.
Both GPUs are running the GPU Driver version 525.125.06 with CUDA SDK 12.0 on Ubuntu 22.04.
PyTorch 2.0.1 is installed on both systems.

\subsection{Training Setup + Metrics}
For evaluation, the following two metrics are used:
The dice coefficient is defined as
\[
Dice = \frac{2 \cdot {TP}}{2 \cdot {TP} + {FP} + {FN}}
\]
As second segmentation metric, we use the Intersection over Union (IoU) which is defined as:
\[
IoU = \frac{|SEG \cap SEG_{GT}|}{|SEG \cup SEG_{GT}|}
\]
For both metrics we use the implementation from MONAI\footnote{https://github.com/Project-MONAI/MONAI} with Cross-entropy loss.
Furthermore, we use the following parameters for training that table~\ref{tab:parameters} shows.
\begin{table*}[]
\begin{center}
\begin{tabular}{ |c|c|} 
 \hline
  \textbf{Parameter} & \textbf{Value}\\ 
  \hline
 Model & \emph{SAM}~\cite{kirillov2023segment}\\ 
 Freeze & Everything except segmentation head\\
 Reduction & Mean\\
 Training Epochs & 75\\
 Steps per Epoch & 75\\
 Learning Rate & 1e-4\\
 Optimizer & Torch Adam\\
 Weight decay & 1e-4\\
 Batch Size & 1\\ 
 Train-/Test Split & 80:20\\
 Split Strategy & Separately by Patient\\
 \hline
\end{tabular}
\caption{Parameters for the SAM Experiments}
\label{tab:parameters}
\end{center}
\end{table*}
\subsection{Sampling strategy}
There are various ways of sampling the input data, which we introduce in the following section:
\subsubsection{Sampling using Positive Points}
\begin{figure}[htbp]
  \centering
  \begin{subfigure}{0.2\textwidth}
    \includegraphics[width=\textwidth]{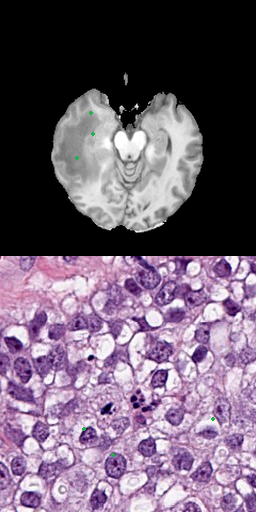}
    \caption{3 positive points}
  \end{subfigure}
  \hfill
  \begin{subfigure}{0.2\textwidth}
    \includegraphics[width=\textwidth]{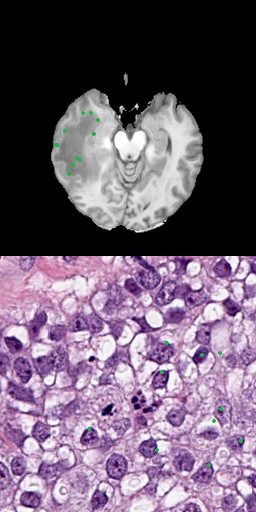}
    \caption{10 positive points}
  \end{subfigure}
  \hfill
  \begin{subfigure}{0.2\textwidth}
    \includegraphics[width=\textwidth]{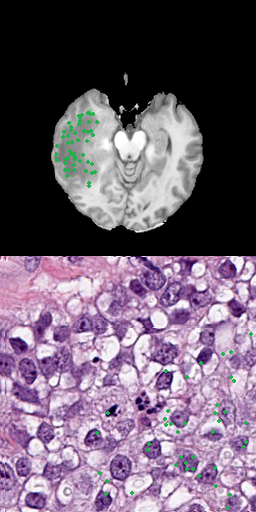}
    \caption{50 positive points}
  \end{subfigure}
  \hfill
  \begin{subfigure}{0.2\textwidth}
    \includegraphics[width=\textwidth]{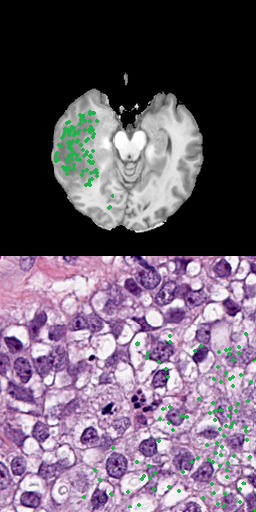}
    \caption{100 positive points}
  \end{subfigure}
  \caption{Results for different number of positive points without bounding box}
  \label{fig:points}
\end{figure}

One approach is to use several \emph{positive points as input data that represent the goal label}~\cite{cheng2023sam}. Using points as input is a precise method, allowing to preserve fine segmentation details but requires more annotation effort. When large areas are annotated, this is especially problematic. Figure~\ref{fig:points} shows samples for using 3, 10, 50 or 100 points for the radiology and pathology data.

\subsubsection{Bounding Box Sampling}
\begin{figure}[htbp]
  \centering
  \begin{subfigure}{0.2\textwidth}
    \includegraphics[width=\textwidth]{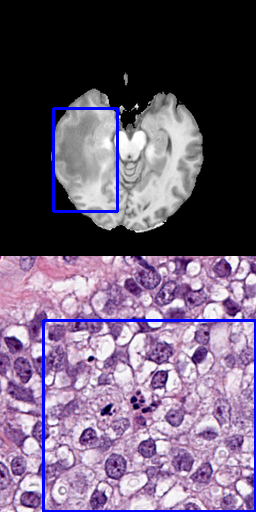}
    \caption{Bounding box without jitter}
  \end{subfigure}
  \hfill
  \begin{subfigure}{0.2\textwidth}
    \includegraphics[width=\textwidth]{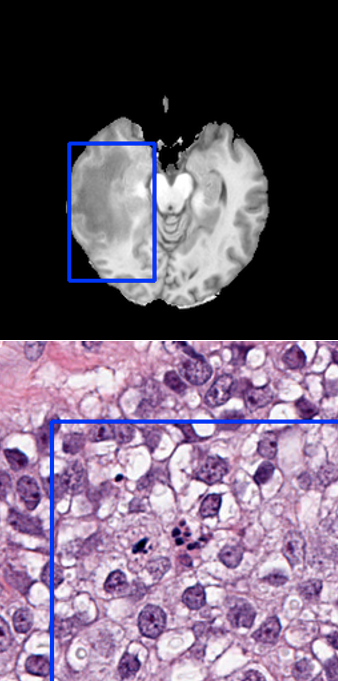}
    \caption{Bounding box with 5\% jitter}
  \end{subfigure}
  \hfill
  \begin{subfigure}{0.2\textwidth}
    \includegraphics[width=\textwidth]{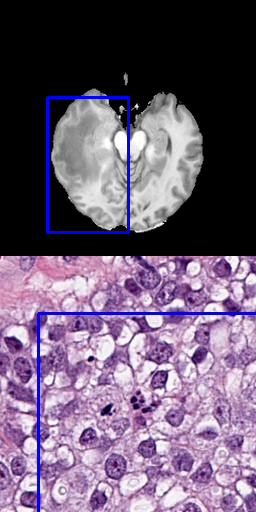}
    \caption{Bounding box with 10\% jitter}
  \end{subfigure}
  \hfill
  \begin{subfigure}{0.2\textwidth}
    \includegraphics[width=\textwidth]{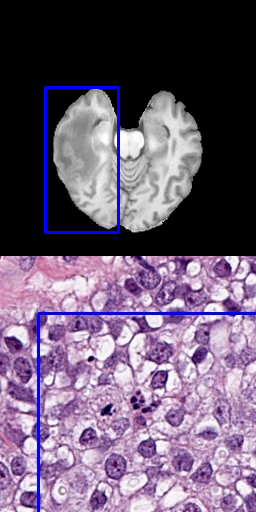}
    \caption{Bounding box with 20\% jitter}
  \end{subfigure}
  \caption{Results for different sampling strategies for the bounding box}
  \label{fig:bb_jitter}
\end{figure}

Alternatively, we evaluate how \emph{bounding boxes}~\cite{cheng2023sam} perform as input. They are easy to annotate while covering larger areas of the image.
On the other hand, they typically can't preserve fine annotation details. We test them as baseline and how adding a certain amount of \emph{jitter}~\cite{cheng2023sam} in all directions changes the segmentation performance.
We compare a bounding box without any jitter with 5\%, 10\% and 20\% jitter. Figure~\ref{fig:bb_jitter} shows samples of such bounding boxes for both domains.

\subsubsection{Bounding Box and Points}
\begin{figure}[htbp]
  \centering
  \begin{subfigure}{0.2\textwidth}
    \includegraphics[width=\textwidth]{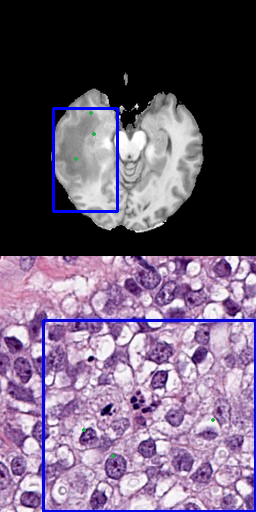}
    \caption{3 points with bounding box}
  \end{subfigure}
  \hfill
  \begin{subfigure}{0.2\textwidth}
    \includegraphics[width=\textwidth]{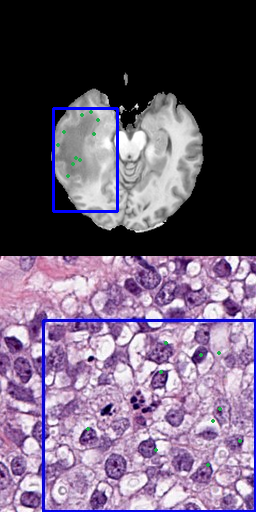}
    \caption{10 points with bounding box}
  \end{subfigure}
  \hfill
  \begin{subfigure}{0.2\textwidth}
    \includegraphics[width=\textwidth]{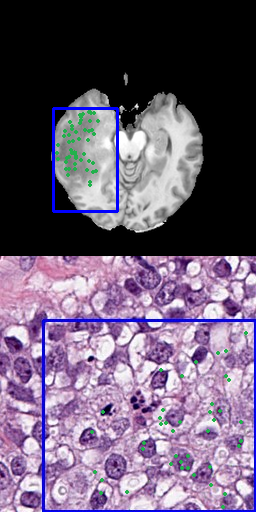}
    \caption{50 points with bounding box}
  \end{subfigure}
  \hfill
  \begin{subfigure}{0.2\textwidth}
    \includegraphics[width=\textwidth]{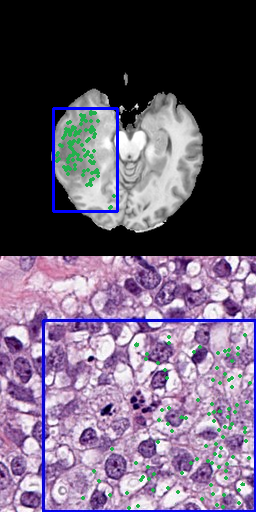}
    \caption{100 points with bounding box}
  \end{subfigure}
  \caption{Results for different number of positive points combined with bounding box}
  \label{fig:points_bb}
\end{figure}

As using only a bounding box is imprecise but requires low effort for large areas and points are more precisely but tedious to set manually, we also tested the combination of both to compensate for their individual disadvantages:
We use 3, 10, 50 and 100 positive input points and combine it with a bounding box that covers a larger area around all of them.
Figure~\ref{fig:points_bb} shows, samples of these for pathology and radiology.

\subsubsection{Special Sampling Strategies}
We evaluate, how \emph{adding negative points}~\cite{cheng2023sam} to the sampling process changes the performance. Therefore, we first test for each domain, which configuration of bounding box and pPoints gives the best segmentation performance. Then we use this configuration but instead replace half of the total points with negative samples. These samples represent the other label than the target label and should provide additional information on the segmentation task.

\begin{figure}[]
  \centering
  \begin{subfigure}{0.2\textwidth}
    \includegraphics[width=\textwidth]{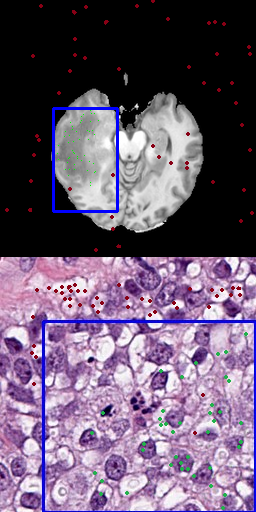}
    \caption{50 positive and 50 negative points}
  \end{subfigure}
  \hfill
  \begin{subfigure}{0.2\textwidth}
    \includegraphics[width=\textwidth]{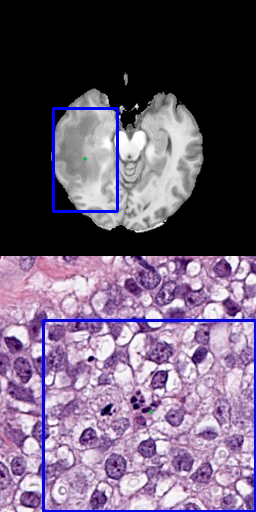}
    \caption{Center of the curated bounding box}
  \end{subfigure}
  \hfill
  \begin{subfigure}{0.2\textwidth}
    \includegraphics[width=\textwidth]{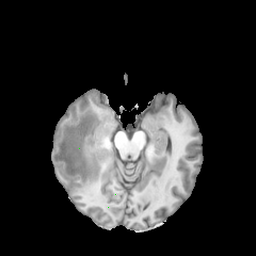}
    \caption{Centroid of the ground truth}
  \end{subfigure}
  \hfill
  \begin{subfigure}{0.2\textwidth}
    \includegraphics[width=\textwidth]{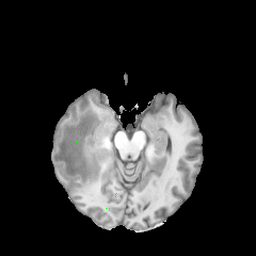}
    \caption{Sampling from 4 ground truth sections}
  \end{subfigure}
  \caption{Results for special sampling strategies}
  \label{fig:special_sampling}
\end{figure}

Moreover, we test, how \emph{using the center point of the bounding box}~\cite{cheng2023sam} influences the segmentation performance.
As an additional sampling method, we test the performance when \emph{using the centroid of the ground truth as input}~\cite{de2023zero}.
As a last variation, we \emph{divide the ground truth into 4 sections and randomly sample one point from each of these sections}~\cite{de2023zero}.
Figure~\ref{fig:special_sampling} shows examples of all these sampling strategies.

\section{Results and Analysis}
\begin{table}[]
\resizebox{\columnwidth}{!}{%
\begin{tabular}{cccccccc}
\hline
\multicolumn{1}{|c|}{\multirow{2}{*}{\textbf{Experiment}}}          & \multicolumn{1}{c|}{\multirow{2}{*}{\textbf{Dataset}}} & \multicolumn{2}{c|}{\textbf{Without BB}}                           & \multicolumn{2}{c|}{\textbf{With BB}}                              & \multicolumn{2}{c|}{\textbf{Improvement with BB (\%)}}               \\ \cline{3-8} 
\multicolumn{1}{|c|}{}                                     & \multicolumn{1}{c|}{}                         & \multicolumn{1}{c|}{\textbf{IoU}}    & \multicolumn{1}{c|}{\textbf{Dice}}   & \multicolumn{1}{c|}{\textbf{IoU}}    & \multicolumn{1}{c|}{\textbf{Dice}}   & \multicolumn{1}{c|}{\textbf{IoU}}      & \multicolumn{1}{c|}{\textbf{Dice}}   \\ \hline
\multicolumn{1}{|c|}{\multirow{2}{*}{3 Positive Points}}   & \multicolumn{1}{c|}{BCSS}                     & \multicolumn{1}{c|}{34.093} & \multicolumn{1}{c|}{39.056} & \multicolumn{1}{c|}{72.311} & \multicolumn{1}{c|}{77.807} & \multicolumn{1}{c|}{+112.099} & \multicolumn{1}{c|}{+99.21} \\ \cline{2-8} 
\multicolumn{1}{|c|}{}                                     & \multicolumn{1}{c|}{BraTS 2020}               & \multicolumn{1}{c|}{61.879} & \multicolumn{1}{c|}{74.217} & \multicolumn{1}{c|}{66.498} & \multicolumn{1}{c|}{77.988} & \multicolumn{1}{c|}{+7.465}   & \multicolumn{1}{c|}{+5.081} \\ \hline
\multicolumn{1}{|c|}{\multirow{2}{*}{10 Positive Points}}  & \multicolumn{1}{c|}{BCSS}                     & \multicolumn{1}{c|}{74.561} & \multicolumn{1}{c|}{79.697} & \multicolumn{1}{c|}{75.331} & \multicolumn{1}{c|}{80.313} & \multicolumn{1}{c|}{+1.033}   & \multicolumn{1}{c|}{+0.773} \\ \cline{2-8} 
\multicolumn{1}{|c|}{}                                     & \multicolumn{1}{c|}{BraTS 2020}               & \multicolumn{1}{c|}{66.818} & \multicolumn{1}{c|}{78.351} & \multicolumn{1}{c|}{68.613} & \multicolumn{1}{c|}{79.738} & \multicolumn{1}{c|}{+2.686}   & \multicolumn{1}{c|}{+1.770} \\ \hline
\multicolumn{1}{|c|}{\multirow{2}{*}{50 Positive Points}}  & \multicolumn{1}{c|}{BCSS}                     & \multicolumn{1}{c|}{79.834} & \multicolumn{1}{c|}{83.511} & \multicolumn{1}{c|}{79.935} & \multicolumn{1}{c|}{83.418} & \multicolumn{1}{c|}{+0.127}   & \multicolumn{1}{c|}{-0.111} \\ \cline{2-8} 
\multicolumn{1}{|c|}{}                                     & \multicolumn{1}{c|}{BraTS 2020}               & \multicolumn{1}{c|}{70.088} & \multicolumn{1}{c|}{80.799} & \multicolumn{1}{c|}{71.715} & \multicolumn{1}{c|}{82.086} & \multicolumn{1}{c|}{+2.321}   & \multicolumn{1}{c|}{+1.593} \\ \hline
\multicolumn{1}{|c|}{\multirow{2}{*}{100 Positive Points}} & \multicolumn{1}{c|}{BCSS}                     & \multicolumn{1}{c|}{\textbf{81.478}} & \multicolumn{1}{c|}{\textbf{84.409}} & \multicolumn{1}{c|}{81.109} & \multicolumn{1}{c|}{84.102} & \multicolumn{1}{c|}{-0.453}   & \multicolumn{1}{c|}{-0.363} \\ \cline{2-8} 
\multicolumn{1}{|c|}{}                                     & \multicolumn{1}{c|}{BraTS 2020}               & \multicolumn{1}{c|}{73.048} & \multicolumn{1}{c|}{82.947} & \multicolumn{1}{c|}{\textbf{73.051}} & \multicolumn{1}{c|}{\textbf{83.029}} & \multicolumn{1}{c|}{+0.004}   & \multicolumn{1}{c|}{+0.099} \\ \hline
\multicolumn{1}{l}{}                                       & \multicolumn{1}{l}{}                          & \multicolumn{1}{l}{}        & \multicolumn{1}{l}{}        & \multicolumn{1}{l}{}        & \multicolumn{1}{l}{}        & \multicolumn{1}{l}{}          & \multicolumn{1}{l}{}        \\
\multicolumn{1}{l}{}                                       & \multicolumn{1}{l}{}                          & \multicolumn{1}{l}{}        & \multicolumn{1}{l}{}        & \multicolumn{1}{l}{}        & \multicolumn{1}{l}{}        & \multicolumn{1}{l}{}          & \multicolumn{1}{l}{}        \\
\multicolumn{1}{l}{}                                       & \multicolumn{1}{l}{}                          & \multicolumn{1}{l}{}        & \multicolumn{1}{l}{}        & \multicolumn{1}{l}{}        & \multicolumn{1}{l}{}        & \multicolumn{1}{l}{}          & \multicolumn{1}{l}{}       
\end{tabular}%
\caption{Intersection over Union (IoU) and Dice for different number of positive points with- and without bounding box}
\label{tab:points}
}
\end{table}

\begin{table}[]
\resizebox{0.7\columnwidth}{!}{%
\begin{tabular}{ccccl}
\cline{1-4}
\multicolumn{1}{|c|}{\multirow{2}{*}{\textbf{Experiment}}}                    & \multicolumn{1}{c|}{\multirow{2}{*}{\textbf{Dataset}}} & \multicolumn{2}{c|}{\textbf{Performance}}                          &  \\ \cline{3-4}
\multicolumn{1}{|c|}{}                                               & \multicolumn{1}{c|}{}                         & \multicolumn{1}{c|}{\textbf{IoU}}    & \multicolumn{1}{c|}{\textbf{Dice}}   &  \\ \cline{1-4}
\multicolumn{1}{|c|}{\multirow{2}{*}{Normal Bounding Box}}           & \multicolumn{1}{c|}{BCSS}                     & \multicolumn{1}{c|}{\textbf{68.673}} & \multicolumn{1}{c|}{\textbf{74.524}} &  \\ \cline{2-4}
\multicolumn{1}{|c|}{}                                               & \multicolumn{1}{c|}{BraTS 2020}               & \multicolumn{1}{c|}{64.406} & \multicolumn{1}{c|}{76.139} &  \\ \cline{1-4}
\multicolumn{1}{|c|}{\multirow{2}{*}{Bounding Box with 5\% Jitter}}  & \multicolumn{1}{c|}{BCSS}                     & \multicolumn{1}{c|}{40.498} & \multicolumn{1}{c|}{46.005} &  \\ \cline{2-4}
\multicolumn{1}{|c|}{}                                               & \multicolumn{1}{c|}{BraTS 2020}               & \multicolumn{1}{c|}{\textbf{64.771}} & \multicolumn{1}{c|}{\textbf{76.421}} &  \\ \cline{1-4}
\multicolumn{1}{|c|}{\multirow{2}{*}{Bounding Box with 10\% Jitter}} & \multicolumn{1}{c|}{BCSS}                     & \multicolumn{1}{c|}{38.927} & \multicolumn{1}{c|}{44.768} &  \\ \cline{2-4}
\multicolumn{1}{|c|}{}                                               & \multicolumn{1}{c|}{BraTS 2020}               & \multicolumn{1}{c|}{64.274} & \multicolumn{1}{c|}{75.884} &  \\ \cline{1-4}
\multicolumn{1}{|c|}{\multirow{2}{*}{Bounding Box with 20\% Jitter}} & \multicolumn{1}{c|}{BCSS}                     & \multicolumn{1}{c|}{39.962} & \multicolumn{1}{c|}{45.577} &  \\ \cline{2-4}
\multicolumn{1}{|c|}{}                                               & \multicolumn{1}{c|}{BraTS 2020}               & \multicolumn{1}{c|}{63.125} & \multicolumn{1}{c|}{75.026} &  \\ \cline{1-4}
\multicolumn{1}{l}{}                                                 & \multicolumn{1}{l}{}                          & \multicolumn{1}{l}{}        & \multicolumn{1}{l}{}        & 
\end{tabular}%
\caption{Intersection over Union (IoU) and Dice for different jitter on the bounding box}
\label{tab:jitter}
}
\end{table}

\begin{table}[]
\resizebox{0.7\columnwidth}{!}{%
\begin{tabular}{ccccl}
\cline{1-4}
\multicolumn{1}{|c|}{\multirow{2}{*}{\textbf{Experiment}}}                            & \multicolumn{1}{c|}{\multirow{2}{*}{\textbf{Dataset}}} & \multicolumn{2}{c|}{\textbf{Performance}}                           &  \\ \cline{3-4}
\multicolumn{1}{|c|}{}                                                       & \multicolumn{1}{c|}{}                         & \multicolumn{1}{c|}{\textbf{IoU}}     & \multicolumn{1}{c|}{\textbf{Dice}}   &  \\ \cline{1-4}
\multicolumn{1}{|c|}{\multirow{2}{*}{50 Positive + 50 Negative Points}}      & \multicolumn{1}{c|}{BCSS}                     & \multicolumn{1}{c|}{34.093}  & \multicolumn{1}{c|}{39.056} &  \\ \cline{2-4}
\multicolumn{1}{|c|}{}                                                       & \multicolumn{1}{c|}{BraTS 2020}               & \multicolumn{1}{c|}{\textbf{74.6944}} & \multicolumn{1}{c|}{\textbf{84.492}} &  \\ \cline{1-4}
\multicolumn{1}{|c|}{\multirow{2}{*}{Center of Curated Bounding Box}}        & \multicolumn{1}{c|}{BCSS}                     & \multicolumn{1}{c|}{\textbf{35.393}}  & \multicolumn{1}{c|}{\textbf{40.532}} &  \\ \cline{2-4}
\multicolumn{1}{|c|}{}                                                       & \multicolumn{1}{c|}{BraTS 2020}               & \multicolumn{1}{c|}{64.157}  & \multicolumn{1}{c|}{75.920} &  \\ \cline{1-4}
\multicolumn{1}{|c|}{\multirow{2}{*}{Centroid of Ground Truth}}              & \multicolumn{1}{c|}{BCSS}                     & \multicolumn{1}{c|}{-}       & \multicolumn{1}{c|}{-}      &  \\ \cline{2-4}
\multicolumn{1}{|c|}{}                                                       & \multicolumn{1}{c|}{BraTS 2020}               & \multicolumn{1}{c|}{50.788}  & \multicolumn{1}{c|}{63.384} &  \\ \cline{1-4}
\multicolumn{1}{|c|}{\multirow{2}{*}{Dividing Ground Truth into 4 Sections}} & \multicolumn{1}{c|}{BCSS}                     & \multicolumn{1}{c|}{-}       & \multicolumn{1}{c|}{-}      &  \\ \cline{2-4}
\multicolumn{1}{|c|}{}                                                       & \multicolumn{1}{c|}{BraTS 2020}               & \multicolumn{1}{c|}{57.038}  & \multicolumn{1}{c|}{70.034} &  \\ \cline{1-4}
\multicolumn{1}{l}{}                                                         & \multicolumn{1}{l}{}                          & \multicolumn{1}{l}{}         & \multicolumn{1}{l}{}        & 
\end{tabular}%
\caption{Intersection over Union (IoU) and Dice for special sampling strategies}
\label{tab:special}
}
\end{table}
In the following section, we investigate the impact of the different SAM configurations on the segmentation performance and computational effort. Furthermore, we discuss these results in the context of the given application in radiology and pathology.
\subsection{SAM Ablations}
We split the experiments into the following three categories and discuss the results:

\subsubsection{Number of Points with/without Bounding Box}
\begin{figure}[htbp]
    \centering
    \includegraphics[width=1.0\textwidth]{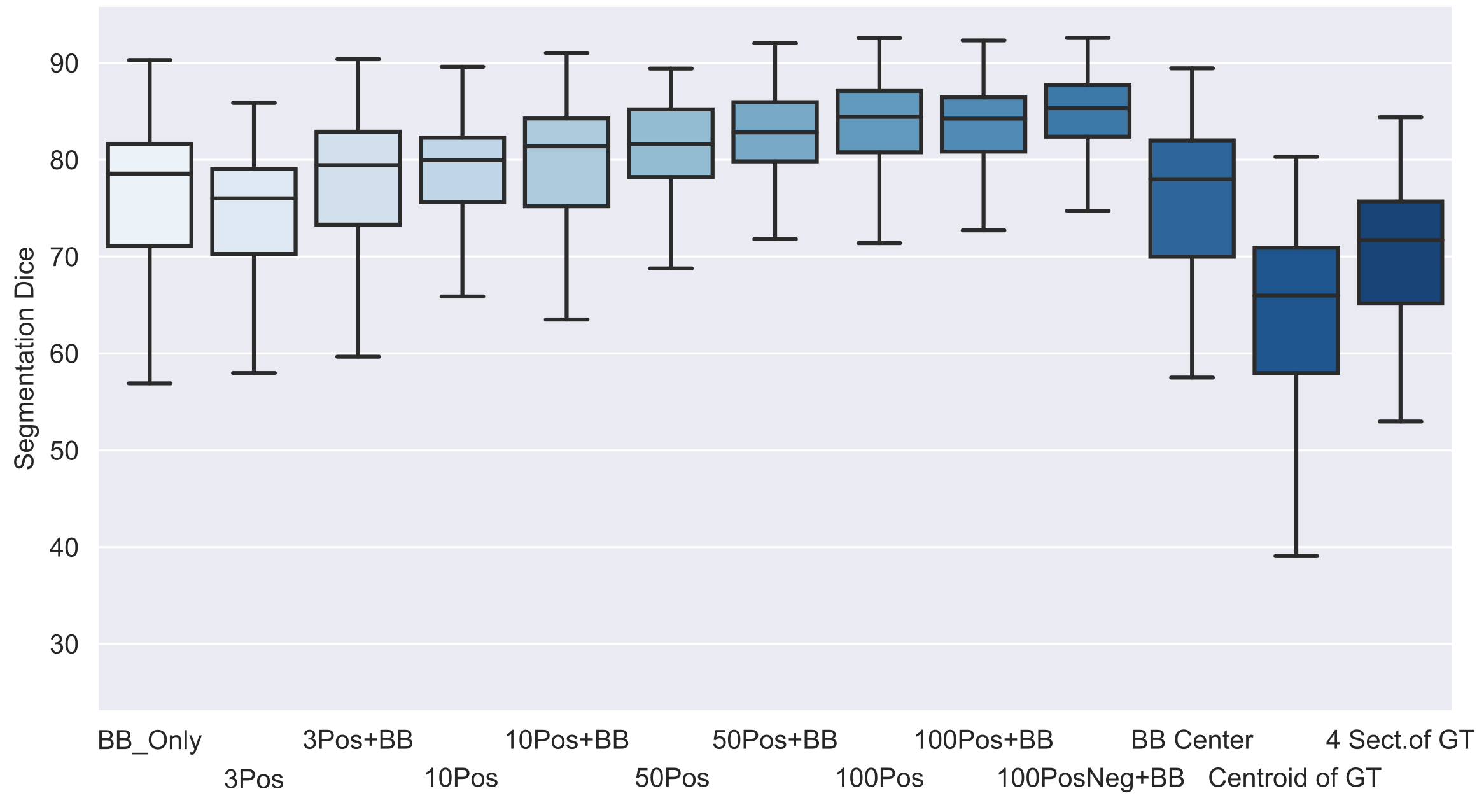}
    \caption{Dice on the radiology dataset BraTS 2020 for the different sampling strategies (number of positive points, with/without bounding box, positive and negative points, curated center of bounding box, centroid of the ground truth and dividing the ground truth in 4 sections and sampling one random point from each)}
    \label{fig:dice_radiology}
\end{figure}
\begin{figure}
    \centering
    \includegraphics[width=1.0\textwidth]{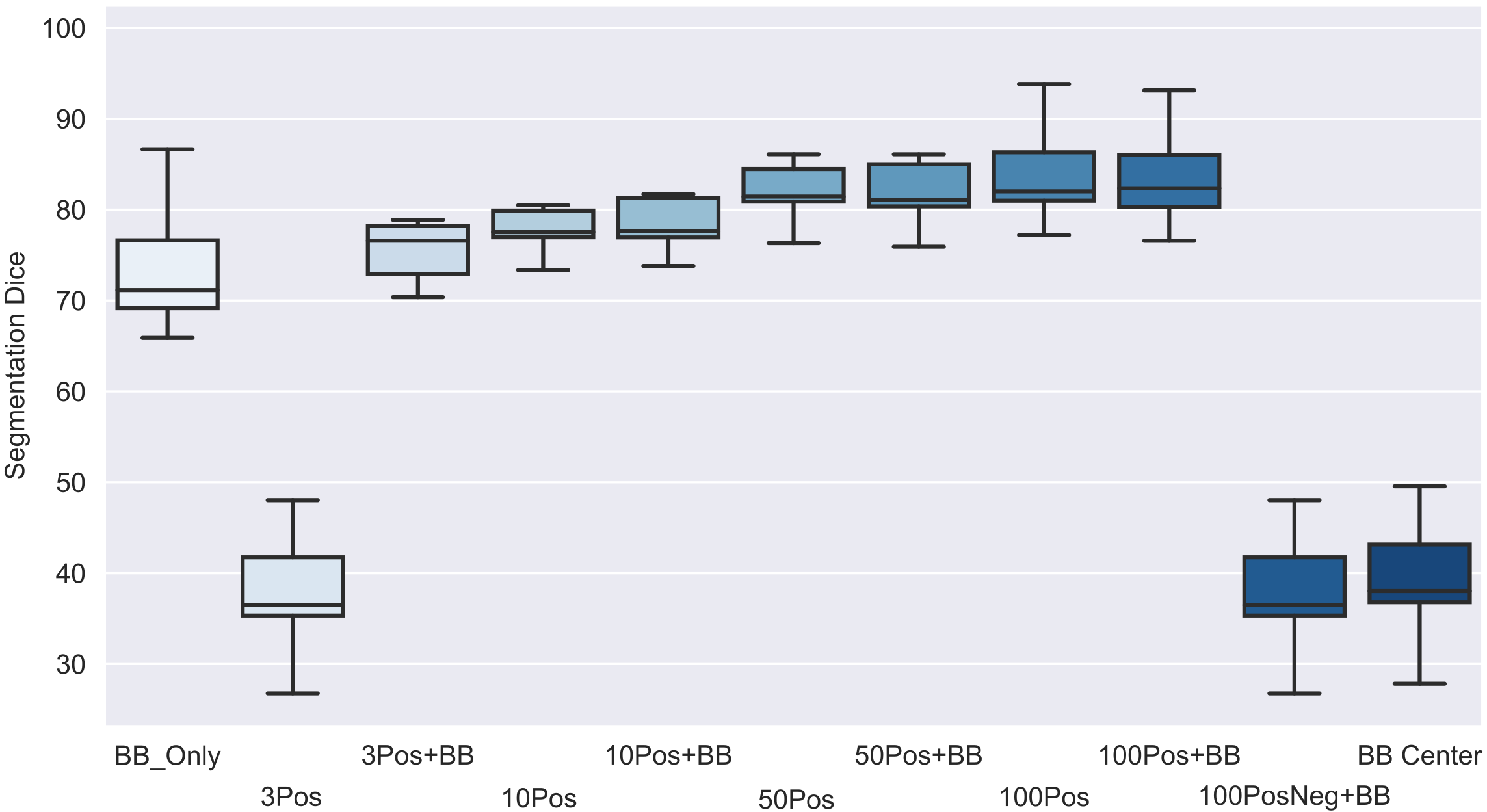}
    \caption{Dice on the pathology dataset BCSS for the different sampling strategies (number of positive points, with/without bounding box, positive and negative points, curated center of bounding box, centroid of the ground truth and dividing the ground truth in 4 sections and sampling one random point from each}
    \label{fig:dice_pathology}
\end{figure}
Table~\ref{tab:points} shows the resulting segmentation performance when applying different numbers of sampling points with- and without bounding boxes.
Figure~\ref{fig:dice_radiology} visualizes them as boxplots for the radiology dataset, while figure~\ref{fig:dice_pathology}  shows the difference of the results in the given pathology dataset.

Both datasets show the trend that more points are consistently improving the segmentation performance. SAM can be trained on more sample points and can generalize better to more complex segmentation tasks. However, BCSS shows a significantly worsened performance with only three sampling points compared to BraTS as the fine segmentation at the cell level is more complex than in MRIs.

Consequently, with only three samples, the performance drops to a Dice of only 39.056\% when using 3 points instead of 10 (Dice 79.697\%). In the case of the radiology domain, this drop of the Dice is only from 66.818\% to 61.879\% due to the simpler features that can be robustly presented with few samples.
Another interesting observation is that while in radiology adding a bounding box always improves the segmentation performance, this doesn't help in the case of pathology and sometimes even decreases the performance. The large bounding box is unable to precisely represent the segmentation at the cell level. In radiology the segmentations are less fine and complex and therefore the bounding box helps to more extensively cover the segmentation which leads to a better model.

The only exception in the case of pathology is when adding a bounding box to the sampling configuration with only 3 points, which improves performance from the Dice of 39.056 to 77.802 as only three points don't cover enough tumor area.

Overall, the results show that in radiology, even only a bounding box and/or three points are sufficient to achieve good performance. On the other hand, in pathology at least 10 points should be given as it significantly improves the results compared to only using 3 points or a bounding box. An alternative would be to use three points but combine it with a bounding box. These results show the potential to reduce workload in clinical applications as SAM is label efficient in both domains, with an advantage in radiology.

\subsubsection{Types of Bounding Boxes}
\begin{figure}[htbp]
  \centering
  \begin{subfigure}{0.49\textwidth}
    \includegraphics[width=\textwidth]{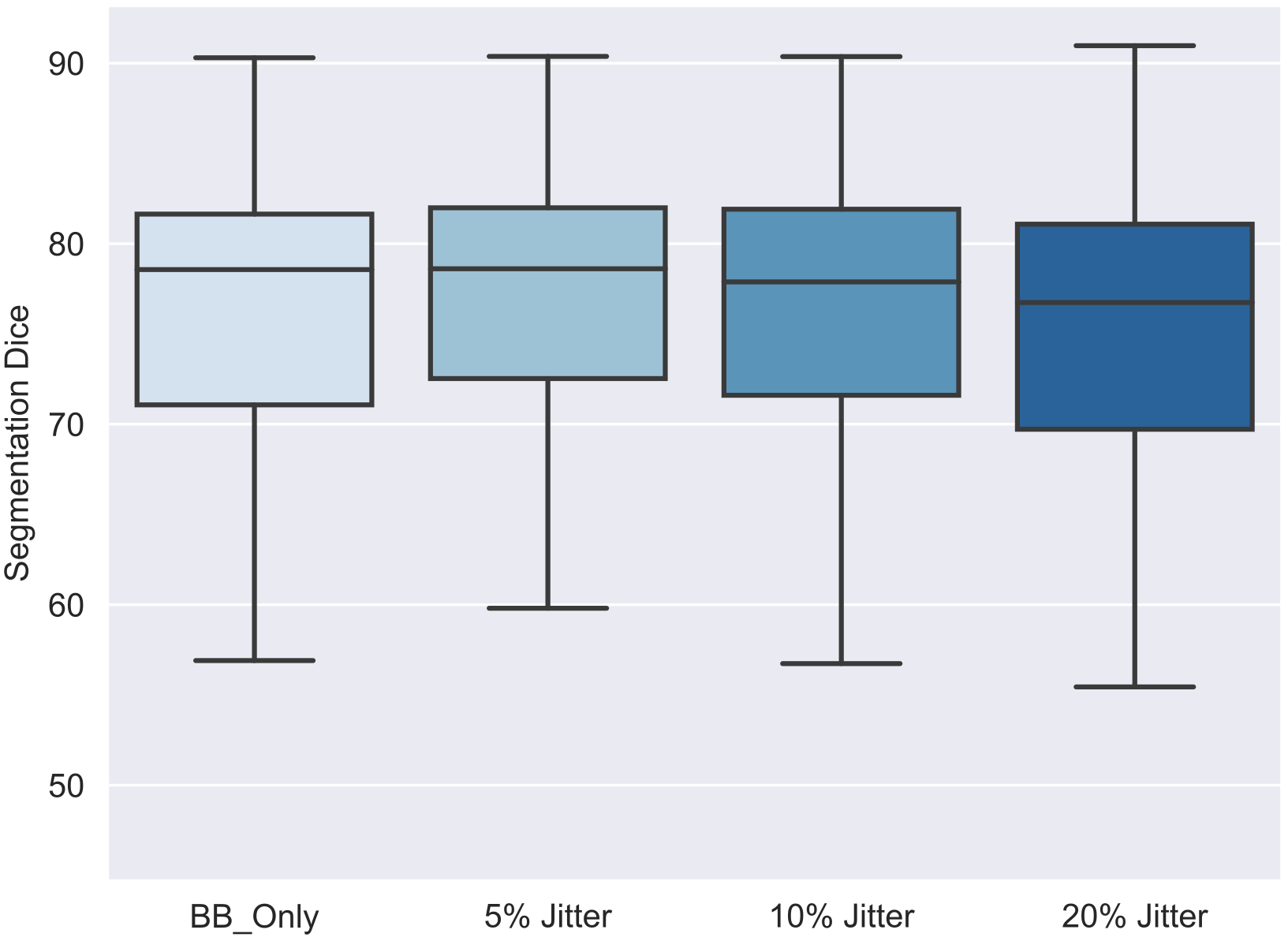}
    \caption{}
    \label{fig:dice_radiology_jitter}
  \end{subfigure}
  \hfill
  \begin{subfigure}{0.49\textwidth}
    \includegraphics[width=\textwidth]{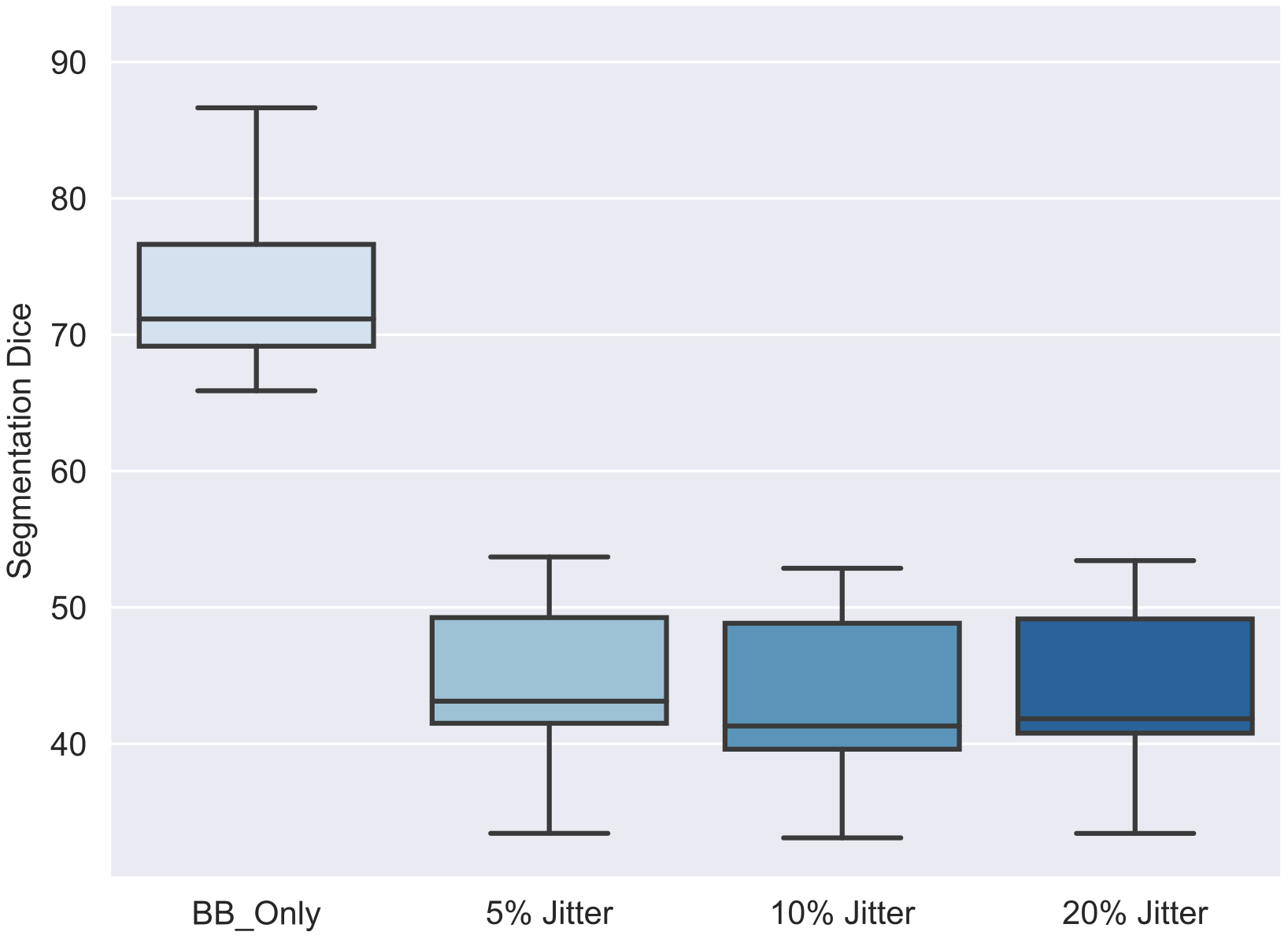}
    \caption{}
    \label{fig:dice_pathology_jitter}
  \end{subfigure}
  \caption{Dice on the radiology dataset BraTS 2020 (a) and the pathology dataset BCSS (b) for different size extensions of the bounding box (5\%, 10\% and 20\% jitter)}
\end{figure}
In additional experiments, we add a jitter of 5\%, 10\% and 20\% to the horizontal and vertical borders of the bounding box, making it larger. Table~\ref{tab:jitter} states the resulting performance during the evaluation of these configurations.

Figure~\ref{fig:dice_radiology_jitter} shows that SAM is robust to the perturbations on radiology data, while figure~\ref{fig:dice_pathology_jitter} shows that the segmentation performance of the SAM model for pathology drops from 74.524\% Dice to 46.005 when adding only 5\% jitter. The fine segmentation by cells instead of MRI areas is more prone to even comparably small changes. Consequently, in a clinical application the bounding boxes should be avoided, if the precise placement of it can't be ensured.

\subsubsection{Special Sampling Strategies}
Table~\ref{tab:special} states the resulting performance during evaluation of the following special configurations.
Figure~\ref{fig:dice_radiology} visualizes them for radiology and figure~\ref{fig:dice_pathology} for pathology.
The VRAM limitation (24 GB) prohibits the use of the sampling strategy of using the centroid of the ground truth as input on BCSS. However, it works on BraTS, but the segmentation performance is the worst of all sampling strategies (Dice 63.384\%) as only one sample is not enough to compete with models trained on more samples.

Equally, when dividing the ground truth into four sections and randomly sampling one point from each, this exceeds the VRAM of the used RTX 4090. The segmentation performance is better than in the previous sampling strategy as more points are present, but still worse than using three or more precisely placed points.

The sampling strategy of using just one point from the center of the bounding box also shows the second-worst performance in radiology as well as pathology.
The performance drop is more severe in pathology, as only one sample is not enough.

As a last sampling strategy, we use the best performing model for each domain with 100 points combined with a bounding box in the case of radiology and no bounding box for pathology and replace half of the positive points with negative points that are used as such by the SAM model.

In pathology, this works well, leading to the best segmentation performance of all sampling strategies. Notably, the performance drops significantly when applying this to the pathology data. We explain this phenomenon with the many labels that are merged in the negative label due to achieving a binary tumor vs. no tumor task in SAM. Too many labels are merged in the negative label, making it problematic to apply negative points.

For the clinical application, this means that using negative points can improve the overall label efficiency in the radiology domain.
In pathology, it's preferable to apply more positive points instead to achieve the best segmentation results.

\subsection{Implementation Effort}
\begin{figure}[h]
    \centering
    \includegraphics[width=1.0\textwidth]{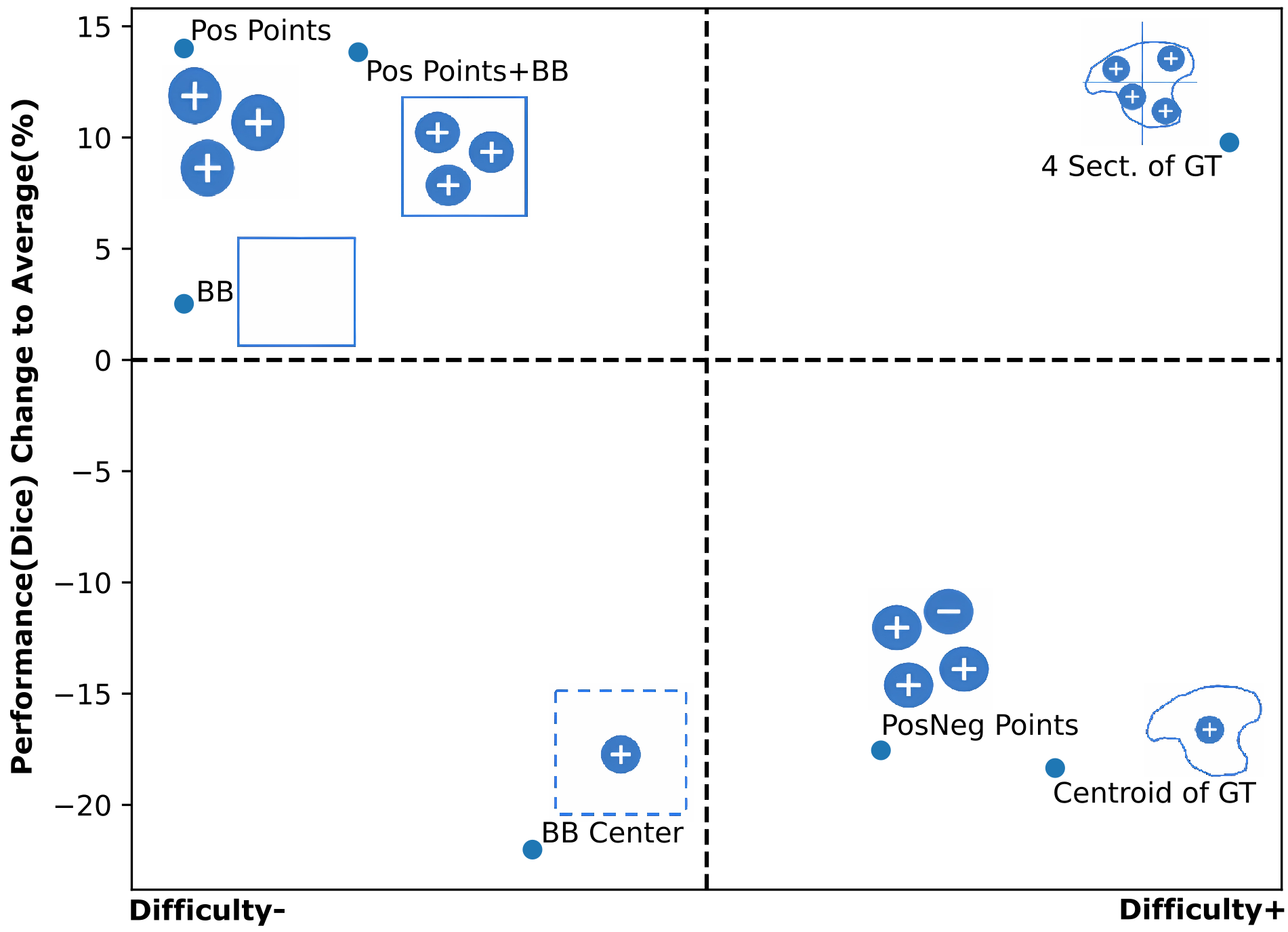}
    \caption{Comparison of the performance improvement and the implementation effort for each evaluated modification. The y-axis represents the performance change to the average while the x-axis indicates the implementation effort. Therefore, the ideal approaches are located top-left.}
    \label{fig:perf_vs_diff}
\end{figure}
To give an overview of which sampling strategy allows the best performance compared to the implementation effort, we visualize the performance difference compared to the average segmentation performance of all approaches to the implementation effort. Figure~\ref{fig:perf_vs_diff} visualizes this as a combined scatter plot for both domains. It shows that only using as many positive points as possible is overall the best way to achieve the best segmentation performance while causing the lowest implementation effort. Alternatively, bounding boxes have a comparably low implementation effort but only perform slightly better than the average. However, if label efficiency with simple annotations is the goal, this brings these advantages.
 \newpage
\section{Conclusion}
    We show that SAM poses great potential for improving medical segmentation tasks.
    Specifically, SAM allows using only a few annotations and our evaluation shows that, especially for simple structures as in Radiology, even providing a few points or just specifying a simple bounding box is sufficient for accurate segmentation results.
    
    In our evaluation with Pathology data, we show that SAM needs more input data than in radiology due to the more complex segmentation at the cell level. However, even here, only as few as 10 positive points bring accurate segmentation potentially preserving patient safety in medical applications.
    This label efficiency of SAM that we observe, allows medical professionals to save time during the labeling process and therefore might enable the use of AI for medical segmentation tasks in more cases.
    
    Furthermore, we evaluate the difficulty of the implementations and show that even the approaches that are the easiest to implement-bounding box and/or positive points-achieve the best segmentation results. This allows to simplify the development for medical applications.
    Nonetheless, the aforementioned discoveries have been made in the fields of pathology and radiology, providing a glimpse into the potential it holds for the entire medical domain.
    
\section{Reproducibility}
All the datasets utilized in this manuscript are accessible to the public via the respective citations. We provide detailed information regarding the trained networks, data partitioning, and instructions for replicating all experiments along with the complete codebase under \url{https://github.com/anon}.

\clearpage
\bibliographystyle{splncs04}
\bibliography{bibliography}
%
% \clearpage

\end{document}